\documentclass[10pt,twocolumn,letterpaper]{article}

\usepackage[pagenumbers]{iccv} % To force page numbers, e.g. for an arXiv version

\definecolor{iccvblue}{rgb}{0.21,0.49,0.74}
\usepackage[pagebackref,breaklinks,colorlinks,allcolors=iccvblue]{hyperref}

\usepackage{url}
\usepackage{svg}
\usepackage{xcolor}
\definecolor{red}{RGB}{255, 0, 0}
\definecolor{blue}{RGB}{0, 0, 255}
\definecolor{magenta}{RGB}{255, 0, 255}
\definecolor{good}{RGB}{54, 183, 0}
\definecolor{bad}{RGB}{157, 44, 0}
\definecolor{comment}{RGB}{159, 43, 104}
\usepackage{xspace}
\usepackage{colortbl}
\usepackage{subcaption}
\usepackage{multirow} 
\usepackage{tikz}
\usepackage{makecell}
\usepackage{algorithm}
\usepackage{algpseudocode} 
\usepackage{amsmath}
\usepackage{listings}
\usepackage{wrapfig}
\newcommand{\tabref}[1]{Table~\ref{#1}}
\newcommand{\figref}[1]{Fig.~\ref{#1}}
\newcommand{\eqnref}[1]{\text{Eq.}~(\ref{#1})}
\newcommand{\secref}[1]{\S\ref{#1}}
\newcommand{\appref}[1]{Appendix~\ref{#1}}
\newcommand{\algname}{\textsc{RankCLIP}\xspace}
\newcommand{\bfsection}[1]{\noindent\textbf{#1}}
\newcommand*\circled[1]{\tikz[baseline=(char.base)]{
            \node[shape=circle,draw,inner sep=2pt] (char) {#1};}}

\def\paperID{11292} % *** Enter the Paper ID here
\def\confName{ICCV}
\def\confYear{2025}

\title{\textsc{\algname}: Ranking-Consistent Language-Image Pretraining}

\author{First Author \\
  Affiliation / Address line 1 \\
  Affiliation / Address line 2 \\
  Affiliation / Address line 3 \\
  \texttt{email@domain} \\\And
  Second Author \\
  Affiliation / Address line 1 \\
  Affiliation / Address line 2 \\
  Affiliation / Address line 3 \\
  \texttt{email@domain} \\}
\author{%
    \textbf{Yiming Zhang}\thanks{These authors contribute equally. Correspondence to Zenghui Ding: dingzenghui@iim.ac.cn}$~^{1,2}$~~~~~\textbf{Zhuokai Zhao}$^{*3}$    \\
    \textbf{Zhaorun Chen$^{3}$~~~Zhili Feng$^{4}$~~~Zenghui Ding$^{1}$~~~Yining Sun$^{1,2}$}  \\
    $^{1}$HFIPS, Chinese Academy of Sciences~~~$^{2}$University of Science and Technology of China\\
    $^{3}$University of Chicago~~~$^{4}$Carnegie Mellon University\\
}
\begin{document}
\maketitle
\vspace{-0.4in}
     
\begin{abstract}
Self-supervised contrastive learning models, such as CLIP, have set new benchmarks for 
vision-language models in many downstream tasks.
However, their dependency on rigid one-to-one mappings overlooks the complex and often
multifaceted relationships between and within texts and images. 
To this end, we introduce \textbf{\algname}, a novel pretraining method that extends 
beyond the rigid one-to-one matching framework of CLIP and its variants. 
By extending the traditional pair-wise loss to list-wise, and leveraging both in-modal and 
cross-modal ranking consistency, \algname improves the alignment process, enabling it to 
capture the nuanced many-to-many relationships between and within each modality.
Through comprehensive experiments, we demonstrate the effectiveness of \algname in 
various downstream tasks, notably achieving significant gains in zero-shot classifications 
over state-of-the-art methods, underscoring the importance of this enhanced learning process.
Code and model checkpoints are available at \url{https://github.com/Jam1ezhang/RankCLIP}.
\end{abstract}
\vspace{-0.2in}
\section{Introduction}\label{sec:introduction}
In the realm of computer vision (CV)~\citep{voulodimos2018deep}, 
natural language processing (NLP)~\citep{chowdhary2020natural}, and multimodal deep 
learning~\citep{jabeen2023review, zhao2023multi, chen2024mj}, the alignment between 
visual and textual modalities~\citep{singh2022flava, chen2024halc} has emerged as a 
cornerstone for downstream applications, ranging from image 
captioning~\citep{ghandi2023deep} to zero-shot 
classification~\citep{pourpanah2022review}.
\begin{figure}
    \centering
    \includegraphics[width=0.5\textwidth]{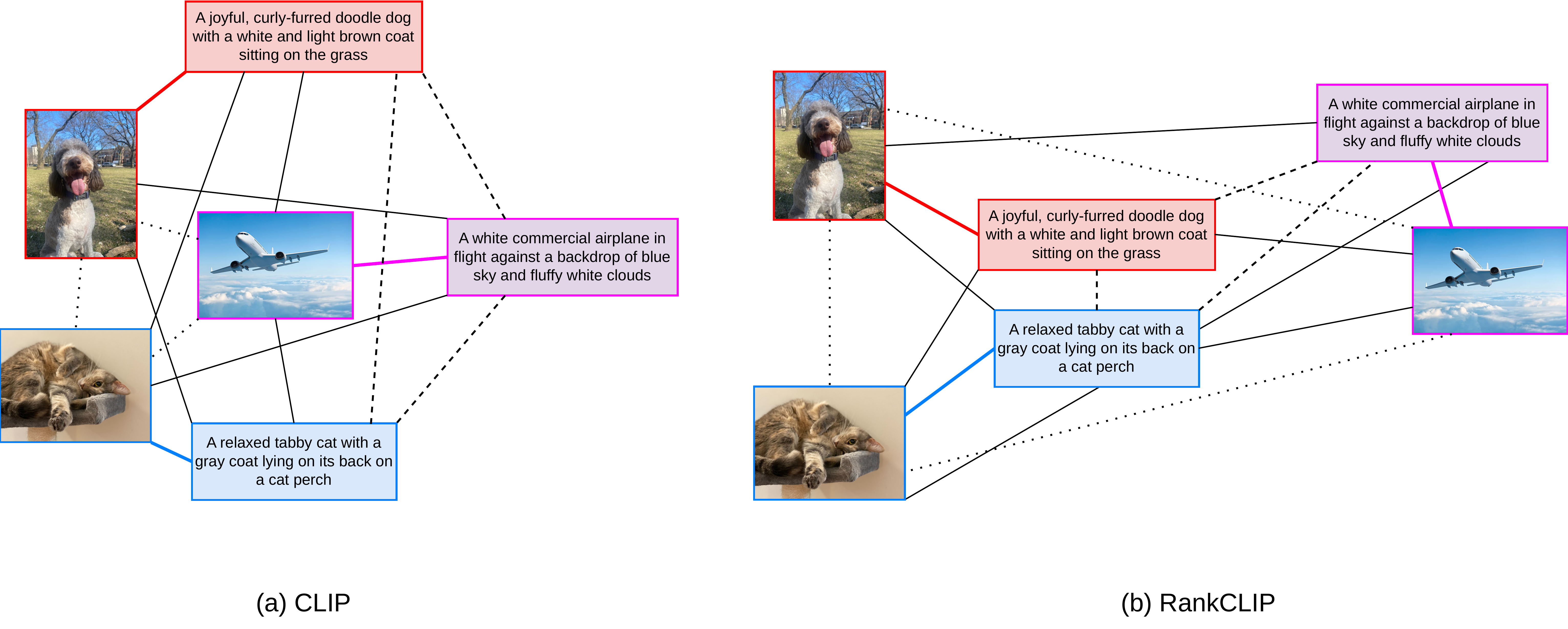}
    \vspace{-0.2in}
    \caption{
        Comparison of learning outcomes between CLIP and \algname using three text-image pairs: \texttt{dog} ({\color{red!30} red}), \texttt{cat} ({\color{blue!30} blue}), and \texttt{car} ({\color{yellow!90} yellow}).
        (a) Contrastive loss treats all unmatched relationships equally, failing to distinguish latent similar attributes between dog and cat versus airplane. 
        \algname addresses this issue by leveraging the shared attributes in (c) during training, improving the final trained embedding distribution from (b) to (d).
    }
    \label{fig:intro_teaser}
    \vspace{-0.25in}
\end{figure}
Contrastive Language-Image Pretraining (CLIP)~\citep{radford2021learning} marks a 
significant advancement in this field, demonstrating incredible performance from training 
on large amounts of text-image pairs to create self-supervised models that 
understand~\citep{hendrycks2021many, hendrycks2021natural, chen2024autoprm}
and generate~\citep{ramesh2021zero, crowson2022vqgan} descriptions of visual contents.
%
% Despite its superior performance, CLIP's reliance on strict one-to-one mappings between 
% images and texts overlooks the nuanced and often many-to-many relationships inherent in 
% the real-world data~\citep{chun2023improved}, leaving rooms for further improvements.
%
Following the success of this contrastive learning paradigm, many recent 
works have been developed upon the original CLIP.
More specifically, these enhancements focus on optimizing data efficiency through 
intrinsic supervision~\citep{li2021supervision}, as well as improving downstream performance 
via cross-modal late interaction~\citep{yao2021filip}, hierarchical feature 
alignment~\citep{gao2022pyramidclip}, geometric consistency 
regularization~\citep{goel2022cyclip}, additional learning~\citep{mu2022slip},
adaptive loss~\citep{yang2023alip}, hierarchy-aware attentions~\citep{geng2023hiclip},
and softer cross-modal alignment~\citep{gao2024softclip}.

Despite the improvements, these methods often have reliance on strict 
\textit{pairwise, cross-modal, and one-to-one} mappings between images and texts, 
overlooking the actual \textit{many-to-many} relationships that exist both \textit{cross-modal}
and \textit{in-modal} in real-world data~\citep{chun2023improved}.
%
% For methods that do concern about relationships between do not fully recognize the 
% many-to-many relationships between and within the image and text modalities, or leverage 
% it to further enhance the model's understanding of complex visual-textual information. 
%
For example, as shown in \figref{fig:intro_teaser}, while pretrained models like CLIP
can correctly classify \texttt{dog}, \texttt{cat} and \texttt{airplane}, they do not 
necessarily learn that \texttt{dog} and \texttt{cat} are more close to each other than 
\texttt{dog} and \texttt{airplane}, in terms of both in-modal (\texttt{dog} text 
is more similar to \texttt{cat} text than to \texttt{airplane} text) and cross-modal 
(\texttt{dog} text is more matched to \texttt{cat} image than to \texttt{airplane} image) 
similarities.
Since it is rooted from the current contrastive loss that only the correct pairs are
optimized while the rest of the unmatched pairs are treated the same, a 
large amount of information not used and unknown to the model during and after the 
training process.

Recognizing the complex \textit{many-to-many} relationships as well as the rich information 
contained within both \textit{in-modal} and \textit{cross-modal} data, we introduce 
\textbf{Rank}ing-\textbf{C}onsistent \textbf{L}anguage-\textbf{I}mage \textbf{P}retraining, 
(\textbf{\algname}), which employs \textit{ranking consistency} to learn and optimize 
similarity levels both between (cross-modal) and within (in-modal) the text-image pairs. 
% in addition to the matched pairs during the self-supervised contrastive training.
%
The concept of ranking consistency stems from the simple observations that similar 
texts often correlate with similar images, as seen with the \texttt{dog}, \texttt{cat} and 
\texttt{airplane} example in \figref{fig:intro_teaser}.
It effectively captures secondary similarity relationships among unmatched pairs, enabling the 
model to learn \textit{more efficiently for free} compared to relying solely on matched pairs.
Ranking consistency is conveniently modeled as an additional loss term to the traditional
contrastive loss, requiring no extra external modules.
It acts as a plug-and-play improvement for many existing methods, including those focusing 
on data-efficiency~\citep{li2021supervision}, potentially boosting performance in both
efficiency and effectiveness.

The main contributions of this paper are: 1) \algname, a novel contrastive 
pretraining method that uses ranking consistency to exploit the 
many-to-many relationships within data, thereby enhancing performance in 
downstream tasks such as zero-shot classification and retrieval accuracy; 
and 2) through comprehensive experiments conducted on multiple datasets, we demonstrate 
the superior effectiveness of \algname in improving pretraining model performance without 
requiring any additional data or computational resources.

\section{Related Work}\label{sec:related_work}
% \subsection{Vision-Language Pretraining}\label{subsec:pretraining}
Vision-language pretraining has witnessed significant advancements over the past 
years~\citep{chen2023vlp, du2022survey, long2022vision}.
Models such as CLIP~\citep{radford2021learning}, ALIGN~\citep{jia2021scaling} and 
FLAVA~\citep{singh2022flava} have pioneered the contrastive learning paradigm applied with 
text-image pairs, showcasing remarkable performance and robustness in downstream tasks.
Many follow-up works, mostly built upon CLIP, have been proposed since then.
% continuing the success of CLIP and its contrastive learning paradigm.
%
\citet{li2021supervision} introduced DeCLIP, improving zero-shot performance through 
intrinsic supervision. 
FILIP~\citep{yao2021filip} advances CLIP's alignment between image patches and text with a 
cross-modal interaction mechanism.
\citet{gao2022pyramidclip} developed PyramidCLIP, using hierarchical feature alignment to 
boost model efficiency and performance. 
Additionally, SLIP~\citep{mu2022slip} merges self-supervised learning with CLIP pre-training 
for improved visual representation and accuracy. 
\citet{goel2022cyclip} introduced CyCLIP, augmenting CLIP with geometric consistency 
regularizers to enhance robustness and performance under varied conditions.
%
% \begin{figure*}[t]
%     \centering
%     \includegraphics[width=.9\textwidth]{figures/overview.pdf}
%     % \vspace{-0.25in}
%     \caption{
%         Illustrative overview of \algname. 
%         %
%         Unlike conventional contrastive loss, which includes only the middle term,
%         \algname introduces both cross-modal and in-modal consistency terms by minimizing
%         a self-supervised, list-wise ranking loss.  
%         %
%         Paired images and texts are indicated by matching contour line colors. 
%         %
%         $V$, $T$, and $S$ represent image embeddings, text embeddings, and similarity scores, 
%         respectively.
%         %
%     }
%     \label{fig:rankclip}
%     \vspace{-0.15in}
% \end{figure*}
% %

Recently, \citet{yang2023alip} introduced ALIP, an adaptive pre-training model that enhances 
language-image alignment using raw text and synthetic captions with dynamic adjustments.
HiCLIP~\citep{geng2023hiclip} refines CLIP by adding hierarchy-aware attentions to uncover 
semantic hierarchies in images and texts. 
EqSim~\citep{wang2023equivariant} incorporates equivariance loss into vision-language 
models, significantly improving sensitivity to semantic changes in image-text pairs.
Additionally, SoftCLIP~\citep{gao2024softclip} softens CLIP’s one-to-one constraint, enabling 
more flexible cross-modal alignment through fine-grained adjustments.

Compared with existing approaches, \algname sets itself apart by fully leveraging the 
\textit{many-to-many} relationships within each batch of text-image pairs, promoting learning 
from both matched and unmatched pairs with varying similarities by integrating in-modal and 
cross-modal \textit{list-wise ranking consistencies} into the contrastive training objective. 
Crucially, \algname diverges from existing models' pair-wise training objective by adopting a 
global, list-wise optimization approach. 
In other words, it considers the rankings of all images and texts collectively within each 
batch, rather than focusing on pairwise similarities as seen in other methods.
%

% \subsection{Learning to Rank}\label{subsec:ranking_loss}
% Among the initial development in learning to rank (LTR) is the pairwise approach, which 
% computes losses based on the relative ordering of item 
% pairs~\citep{burges2005learning, joachims2002optimizing, liu2009learning}.
% %
% Despite the computational efficiency and scalability of pairwise losses, they fall short by
% not accounting for the global ranking context, often leading to suboptimal ranking 
% outcomes~\citep{liu2009learning, burges2010ranknet}.
% %
% To address these limitations, list-wise approaches such as ListNet~\citep{cao2007learning} 
% and ListMLE~\citep{xia2008listwise} were proposed, focusing on optimizing the entire 
% ranking sequence instead.
% %
% More specifically, these strategies employ Plackett-Luce (PL) ranking 
% models~\citep{luce2005individual, plackett1975analysis} to enhance the likelihood of 
% achieving the most accurate item ordering.
% %
% In \algname, we incorporate ListMLE~\citep{cao2007learning} as part of our 
% training objective to optimize both in-modal and cross-modal ranking consistencies.
% %

\section{\algname}\label{sec:rankclip}
\begin{figure*}[t]
    \centering
    \includegraphics[width=.9\textwidth]{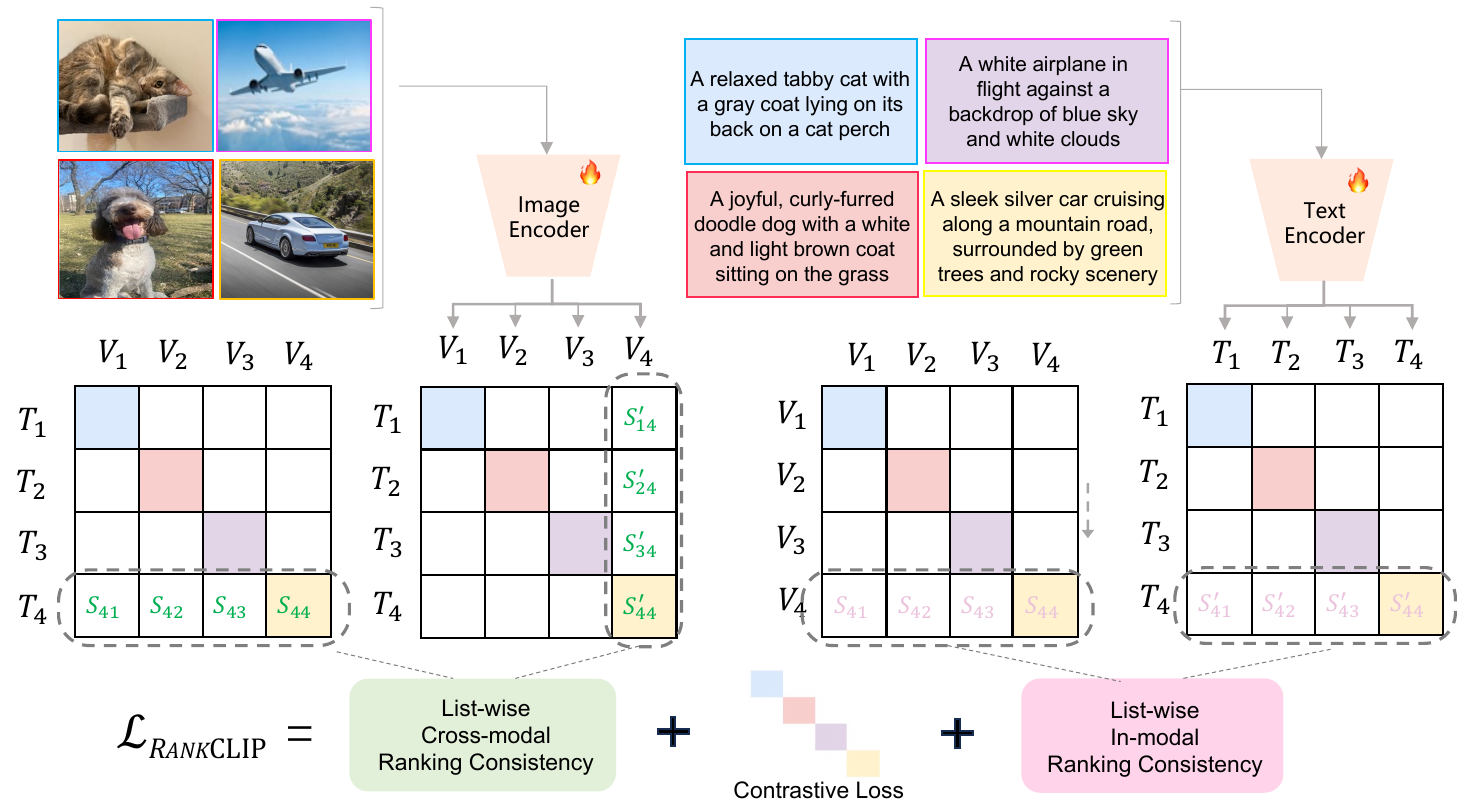}
    % \vspace{-0.25in}
    \caption{
        Overview of \algname. 
        Unlike conventional contrastive loss, which includes only the middle term,
        \algname introduces both cross-modal and in-modal consistency terms by minimizing
        a self-supervised, list-wise ranking loss.  
        Paired images and texts are indicated by matching contour line colors. 
        $V$, $T$, and $S$ represent image embeddings, text embeddings, and similarity scores, 
        respectively.
    }
    \label{fig:rankclip}
    \vspace{-0.15in}
\end{figure*}
\algname efficiently leverages the many-to-many relationships in real-world data by focusing on 
both matched and unmatched pairs. 
As in \figref{fig:rankclip}, it not only identifies if an image-text pair matches but 
also assesses their relative semantic similarities to other images and texts of both 
modalities in the dataset through self-supervised ranking consistency. 
Uniquely, \algname employs a list-wise loss for training batches, distinguishing it from other 
methods that solely rely on pair-wise relationships, as discussed in \secref{sec:related_work}.

\subsection{Ranking Model Formulation}\label{subsec:ranking_formulation}
\algname leverages the Plackett-Luce (PL) ranking
model~\cite{plackett1975analysis, luce2005individual, guiver2009bayesian} to estimate the 
probability distribution over rankings for every image-text pair $(V_i, T_j)$, so that the 
consistency in their relative ordering with respect to a reference ranking can be measured.
Specifically, for a given data pair, whether it is in-modal (image-image, text-text), or
cross-modal (image-text), we calculate its in- or cross-modal cosine similarity $S_{ij}$ to 
serve as the score when measuring the alignment of its ranking with respect to another 
reference ranking $y_{\text{ref}}$.

Following~\cite{plackett1975analysis}, we first sort the reference ranking in a descending 
order to construct the optimal ranking $y^*$, and assume that the ego ranking $y$ is sampled 
from $y^*$. 
The probability that item $d$ with score $S_{ij}$ is ranked $k^{\text{th}}$ in the ego 
ranking $y$ from a set of items $\mathcal{D}$ is the score of $e^{S_{ij}}$ divided by the sum 
of scores for the items that have not been placed yet:
\begin{equation}
\pi(d \mid y_{1:k-1}, y_{\text{ref}}, \mathcal{D}) = \frac{e^{S_{ij}}}{\sum_{d' \in \mathcal{D} \setminus y_{1:k-1}} e^{S'_{ij}}},
\end{equation}
where $y_{1:k-1} = [y_1, y_2, ..., y_{k-1}]$ denotes the set of items ranked before $d$.
%
% In addition, we propose a decaying factor $\mu = 1/\log(k + 1)$ to scale the loss, so that the 
% top-ranked items can obtain higher weights:
%
Consequently, the probability of the entire ranking $y$ is the product of all individual 
placement probabilities:
\begin{equation}
\mathcal{P}(y, y_{\text{ref}}) = \prod_{k=1}^{K} \pi(y_k \mid y_{1:k-1}, \mathbf{y_{\text{ref}}}, \mathcal{D}).
\end{equation}
\algname's objective is to maximize the consistency log-likelihood of the list ranking in 
one modality towards the reference ranking (from the same/in-modal and different/cross-modal
data), which conveniently aligns with minimizing the negative log-likelihood loss:
\begin{equation}\label{eqn:pl_loss}
\mathcal{L}_{\text{PL}} = -\log \mathcal{P}(y, y_{\text{ref}})
\end{equation}
%where $y$ and $y_{\text{ref}}$ can be in either modality. 
%

\subsection{Cross-modal Consistency Ranking}\label{subsec:cross-modal}
As illustrated by the green box in \figref{fig:rankclip}, \algname utilizes secondary 
relationships between unmatched visual and textual representations by constructing a 
list-wise rank loss.
This approach ensures that the semantic similarity rankings between one image and multiple
texts align with those between one corresponding text and multiple images.
For example, as shown in \figref{fig:intro_teaser}, from the \texttt{dog} perspective, 
the semantic distance between \texttt{dog} image and \texttt{cat} text is closer compared to
the \texttt{plane} text.
This relationship should also apply between the \texttt{dog} text and the \texttt{cat}, 
\texttt{plane} images.
Mathematically, \eqnref{eqn:pl_loss} can be specified as:
\begin{align}\label{eqn:cross-modal loss}
   \mathcal{L}_{\text{cross-modal}} & = -\log \mathcal{P}(\bf{y}_{\text{image-text}}, \bf{y}_{\text{text-image}}) \\
   & = -\log \mathcal{P}(\bf{\hat{v}} \cdot \bf{\hat{t}}^T, \bf{\hat{t}} \cdot \bf{\hat{v}}^T)
\end{align}
By optimizing \eqnref{eqn:cross-modal loss}, \algname enhances its ability to bridge the 
semantic gap between modalities by leveraging nuanced unmatched correlations. 
This can also be viewed as learning a symmetric cosine-similarity matrix, further reinforcing 
semantic consistency across modalities.
%
% We can also interpret \eqnref{eqn:cross-modal loss} as learning a \textit{symmetric} 
% cosine-similarity matrix to further enforce the semantic consistency between both modalities.
%
\begin{table*}[t]
\centering

\vspace{-0.1in}
\setlength{\tabcolsep}{11pt}
\renewcommand{\arraystretch}{1.2}
\resizebox{\linewidth}{!}{
\begin{tabular}{c | c c c | c c c | c c c}
\hline
& \multicolumn{3}{c|}{\multirow{2}{*}{ImageNet1K}} & \multicolumn{6}{c}{MSCOCO}   \\
\cline{5-10}
&    &    &    & \multicolumn{3}{c|}{Image-to-Text Retrieval} & \multicolumn{3}{c}{Text-to-Image Retrieval}    \\
\cline{2-10}
& Top-1 & Top-3 & Top-5 & Recall@1 & Recall@5 & Recall@10 & Recall@1 & Recall@5 & Recall@10  \\
\hline
CLIP~\citep{radford2021learning} & 9.06\% & 16.94\% & 21.63\% & 6.68\% & 18.36\% & 26.94\%  & 3.70\% & 9.74\% & 14.04\% \\
% \hline
CyCLIP~\citep{goel2022cyclip} & 9.40\% & 17.32\% & 21.72\% & 6.50\% & 19.34\% & \textbf{29.14\%}  & 3.72\% & \textbf{11.16\%} & \textbf{16.06\%} \\

ALIP~\citep{yang2023alip} & 9.71\% & 18.31\% & 23.07\% & 6.04\% & 18.04\% & 26.92\% & 3.70\% & 10.22\% & 14.38\%   \\
\hline
\rowcolor{gray!20} \algname 
&\textbf{10.16\%} &\textbf{19.57\%} &\textbf{24.01\%} &\textbf{7.18\%} &\textbf{19.46\%} &28.48\% &\textbf{3.74\%} &10.28\% &14.18\% \\
\hline
\end{tabular}
}
\caption{
    Zero-shot top-1, top-3, and top-5 classification accuracy on ImageNet1K, along with retrieval performance on MS-COCO. 
    The proposed \algname consistently outperforms all baselines across both tasks. 
    All models are trained on CC3M with ViT-B/32 backbone.
}
\label{tab:zero_shot_results}
\vspace{-0.15in}
\end{table*}
%
% & \makecell{\textbf{10.16\%} \\ \textbf{({\color{good}+1.1\%})}} 
% & \makecell{\textbf{19.57\%} \\ ({\color{good}+2.63\%})}
% & \makecell{\textbf{24.01\%} \\ ({\color{good}+2.38\%})}
% & \makecell{\textbf{7.18\%} \\ \textbf{({\color{good}+0.5\%})}} 
% & \makecell{\textbf{19.46\%} \\ \textbf{({\color{good}+2.77\%})}} 
% & \makecell{\textbf{28.48\%} \\ \textbf{({\color{good}+1.1\%})}}  
% & \makecell{\textbf{3.74\%} \\ \textbf{({\color{good}+0.04\%})}} 
% & \makecell{\textbf{10.28\%} \\ \textbf{({\color{good}+0.54\%})}} 
% & \makecell{\textbf{14.18\%} \\ \textbf{({\color{good}+0.14\%})}}  \\
%

\subsection{In-modal Consistency Ranking}\label{subsec:in-modal}
The pink box in \figref{fig:rankclip} highlights the in-modal consistency component of the
proposed rank loss.
\algname ensures semantic consistency within each modality -- image to image and text to text
-- enhancing the use of secondary unmatched relationships as an optimization objective.
The underlying principle is that similar images should correspond to similar texts.
For example, in \figref{fig:intro_teaser}, from the \texttt{dog} image perspective, the
\texttt{cat} image is the most similar, followed by the \texttt{plane} image. 
This relationship should hold true for their corresponding texts as well, where we utilize this
to construct our $y$ and $y_{\text{ref}}$ from \eqnref{eqn:pl_loss}.
Mathematically, \eqnref{eqn:pl_loss} can be specified as:
\begin{align}\label{eqn:in-modal_loss}
   \mathcal{L}_{\text{in-modal}} & = -\log \mathcal{P}(\bf{y}_{\text{text-text}}, \bf{y}_{\text{image-image}}) \\
   & = -\log \mathcal{P}(\bf{\hat{t}} \cdot \bf{\hat{t}}^T, \bf{\hat{v}} \cdot \bf{\hat{v}}^T)
\end{align}
where $\bf{\hat{t}}$ and $\bf{\hat{v}}$ are the text and image batch embedding matrix, 
respectively. 
Via \eqnref{eqn:in-modal_loss}, the model can efficiently leverage the nuanced in-modal 
relationships to learn a richer and more structured semantic representation.

\subsection{\algname Loss}\label{subsec:rankclip_loss}
Combining both cross-modal and in-modal consistency with the traditional contrastive loss 
(more details in \appref{app:clip}), the complete rank loss is thus formulated as:
\begin{equation}\label{eqn:rankclip_loss}
   \mathcal{L}_{\text{\algname}} = \mathcal{L}_{\text{CLIP}} + \lambda_1 \mathcal{L}_{\text{in-modal}} + \lambda_2 \mathcal{L}_{\text{cross-modal}}
\end{equation}
which is also depicted in \figref{fig:rankclip}.
By supplementing the pairwise contrastive loss with cross-modal and in-modality ranking 
consistency loss, \algname systematically organizes embeddings to fully exploit both global 
and fine-grained unmatched relationships, which enhances the learning of more 
informative and accurate representations, better supporting downstream multi-modal tasks.

\subsection{Training Recipe on Selecting $\lambda_1$ and $\lambda_2$}\label{subsec:training_recipe}
In the early stage of pre-training, rank consistency is highly unstable due to random initialization.
Overemphasizing ranking consistency at this stage can impede the optimization of the embedding space.
To address this, we gradually increase the weights ${\lambda}_1$ and ${\lambda}_2$ of the ranking loss as training progresses.
Specifically, we have:
\begin{equation}
    \lambda_1 = \lambda_2 = \text{clip}\left(\frac{3i - 1}{n - 1}, 0, 2\right) \notag
\end{equation}
where $i$ and $n$ denote the current training epoch and total number of epoch, respectively.
The full \algname framework is outlined in Algorithm~\ref{alg:pseudocode}.

\section{Experiments}\label{sec:experiments}
\subsection{Experimental Setup}\label{subsec:setup}

\bfsection{Baselines.}  
The most direct baseline to \algname is the original CLIP~\citep{radford2021learning}.  
To further demonstrate the superior performance of \algname, we include CyCLIP~\citep{goel2022cyclip}, which introduces cyclic consistency constraints to enforce more robust alignment between visual and textual representations, improving generalization and semantic coherence.  
We also include ALIP~\citep{yang2023alip}, which leverages synthetic captions to enhance vision-language representation learning. 
More specifically, it employs a unique architecture that dynamically adjusts sample and pair weights to mitigate the impact of noisy or irrelevant data, making its approach complementary to ours.  
The training procedures and parameters for all models are detailed in \appref{app:training_procedure}.

\bfsection{Data.}
All approaches are pretrained on the Conceptual Captions 3M (CC3M) dataset~\citep{sharma2018conceptual}, which contains approximately 3.3 million text-image pairs.  
Although significantly smaller than CLIP’s original dataset~\citep{ilharco_gabriel_2021_5143773}, CC3M remains a standard benchmark for vision-language pretraining, enabling strong zero-shot performance~\citep{carlini2021poisoning, li2021supervision, tejankar2021fistful, mu2022slip, goel2022cyclip}.  
As discussed in \secref{subsec:scaling}, we also train CLIP and \algname on a larger 15M-image subset of YFCC100M~\citep{Thomee2016YFCC100M} (YFCC15M) to study the effect of scaling up the dataset size.
\begin{table*}[t]
\centering

\vspace{-0.1in}
\setlength{\tabcolsep}{4pt}
\renewcommand{\arraystretch}{1.2}
\resizebox{\linewidth}{!}{
\begin{tabular}{c | c c c | c c c | c c c | c c c}
\hline
& \multicolumn{3}{c|}{ImageNetV2-Matched} &  \multicolumn{3}{c|}{ImageNetV2-Threshold} &  \multicolumn{3}{c|}{ImageNetV2-Top} & \multicolumn{3}{c}{ImageNet-R} \\
\hline
& Top1 & Top3 & Top5 & Top1 & Top3 & Top5 & Top1 & Top3 & Top5 & Top1 & Top3 & Top5 \\
\hline
CLIP~\citep{radford2021learning} & 7.53\% & 14.99\% & 19.61\% & 8.89\% & 17.22\% & 21.86\% & 10.76\% & 19.80\% & 24.87\% & 9.36\% & 10.56\% & 19.76\%\\

CyCLIP~\citep{goel2022cyclip} & 7.68\% & 15.07\% & 19.11\% & 9.10\% & 17.42\% & 21.94\% & 11.20\% & 20.18\% & 25.34\% & 9.23\% & 16.72\% & 21.64\% \\
% \hline
ALIP~\citep{yang2023alip} & 7.82\% & 15.56\% & 19.81\% & 9.65\% & 18.31\% & 22.85\% & 11.43\% & 20.88\% & 26.10\% & 10.92\% & 20.27\% & 26.24\% \\
\hline
\rowcolor{gray!20} \algname
& \makecell{\textbf{9.01\%}}  
% \textbf{({\color{good}+1.19\%})}}
& \makecell{\textbf{16.95\%}}  
% \textbf{({\color{good}+1.39\%})}}
& \makecell{\textbf{21.12\%}}  
% \textbf{({\color{good}+1.31\%})}}
& \makecell{\textbf{10.32\%}}  
% \textbf{({\color{good}+0.67\%})}}
& \makecell{\textbf{19.31\%}}  
% \textbf{({\color{good}+1.00\%})}}
& \makecell{\textbf{24.13\%}}  
% \textbf{({\color{good}+1.28\%})}}
& \makecell{\textbf{12.31\%}}  
% \textbf{({\color{good}+0.88\%})}}
& \makecell{\textbf{22.11\%}}  
% \textbf{({\color{good}+1.23\%})}}
& \makecell{\textbf{27.17\%}}  
% \textbf{({\color{good}+1.07\%})}}
& \makecell{\textbf{11.34\%}}  
% \textbf{({\color{good}+1.98\%})}}
& \makecell{\textbf{20.88\%}}
% \textbf{({\color{good}+10.32\%})}}
& \makecell{\textbf{26.94\%}} \\ 
% \textbf{({\color{good}+7.18\%})}}  \\
\hline
\end{tabular}
}
\caption{
    Zero-shot top-1, 3, and 5 accuracy on ImageNet1K variants with \textit{natural distribution shifts}.  
    Compared to baselines, \algname achieves higher accuracies.
    Notably, these gains are more pronounced than on standard ImageNet1K, highlighting improved robustness.  
}
\label{tab:zero_shot_shifted_results}
\vspace{-0.1in}
\end{table*}
\begin{table*}[t]
\centering
% \vspace{-0.1in}
\setlength{\tabcolsep}{8pt}
\renewcommand{\arraystretch}{1.2}
\resizebox{\linewidth}{!}{
\begin{tabular}{c | c c c c c c c c c c | c }
\hline
& \rotatebox{55}{CIFAR-10} & \rotatebox{55}{CIFAR-100} & \rotatebox{55}{DTD} & \rotatebox{55}{FGVGAircraft} & \rotatebox{55}{Food101} & \rotatebox{55}{GTSRB}  & \rotatebox{55}{OxfordPets} & \rotatebox{55}{SST2} & \rotatebox{55}{STL10} & \rotatebox{55}{SVHN} & \rotatebox{55}{Average} \\
\hline
CLIP~\citep{radford2021learning} & 77.6\% & 56.2\% & 43.2\% & 22.6\% & 39.7\% & 60.0\%  & 40.4\% & 51.0\% & 79.0\% & \textbf{50.5\%} & 52.0\% \\
% \hline
CyCLIP~\citep{goel2022cyclip} & 76.8\% & 54.3\% & 45.8\% & 19.2\% & 37.5\% & 58.6\% & \textbf{44.2\%} & 51.5\% & \textbf{82.3\%} & 41.3\% & 51.2\% \\
ALIP~\citep{yang2023alip} & 71.1\% & 49.1\% & \textbf{47.1\%} & 17.4\% & 36.1\% & 51.5\% & 41.9\% & 53.3\% & 81.0\% & 38.3\% & 48.7\% \\
\hline
\rowcolor{gray!20}  \algname & \textbf{78.4\%} & \textbf{56.6\%} & 42.4\% & \textbf{23.4\%}  & \textbf{40.2\%} & \textbf{60.6\%} & 40.6\%  & \textbf{53.4\%}  & 79.6\%  & 47.7\%  & \textbf{52.3\%} \\

% \algname
% & \makecell{78.4\%} %\\ \textbf{({\color{good}+0.82\%})}}
% & \makecell{56.6\%} %\\ \textbf{({\color{good}+0.49\%})}}
% & \makecell{42.4\%} %\\ ({\color{bad}-0.80\%})}
% & \makecell{23.4\%} %\\ \textbf{({\color{good}+0.84\%})}}
% & \makecell{40.2\%} %\\ \textbf{({\color{good}+0.47\%})}}
% & \makecell{60.6\%} %\\ ({\color{bad}-1.42\%})}
% & \makecell{40.6\%} %\\ \textbf{({\color{good}+0.17\%})}}
% & \makecell{53.4\%} %\\ \textbf{({\color{good}+2.36\%})}}
% & \makecell{79.6\%} %\\ \textbf{({\color{good}+0.61\%})}}
% & \makecell{47.7\%} %\\ ({\color{bad}-2.81\%})}
% & \makecell{\textbf{52.23\%}} %\\ \textbf{({\color{good}+0.29\%})}} \\
% \\
\hline
\end{tabular}
}
\caption{
    Linear probing accuracy on 10 downstream datasets using a ViT backbone. 
}
\label{tab:linear_probe_results}
\vspace{-0.1in}
\end{table*}

\subsection{Zero-shot Classification}\label{subsec:zero_shot}
We evaluate the zero-shot classification performance of CLIP~\citep{radford2021learning}, CyCLIP~\citep{goel2022cyclip}, ALIP~\citep{yang2023alip}, and \algname on ImageNet1K~\citep{deng2009imagenet, imagenet15russakovsky}.
As shown in \tabref{tab:zero_shot_results}, \algname consistently outperforms CLIP, which highlight the effectiveness of ranking consistency in enhancing language-image alignment with the same training data.  
Compared to CyCLIP~\citep{goel2022cyclip}, which enforces cyclic consistency to improve semantic coherence, \algname achieves higher accuracy across all metrics, suggesting that ranking consistency provides a more direct and effective regularization for representation learning.
Additionally, \algname surpasses ALIP~\citep{yang2023alip}, indicating that ranking consistency is a stronger alternative to synthetic caption-based supervision.
Notably, \algname shows the most significant improvement in top-1 accuracy, reinforcing its practical advantages where the highest-ranked prediction is most critical.  

\subsection{Zero-shot Cross-modal Retrieval}\label{subsec:image_text_retrieval}
We further evaluate \algname on zero-shot cross-modal retrieval tasks, including image-to-text and text-to-image retrieval, using the MSCOCO~\citep{lin2014microsoft} dataset.  
As shown in \tabref{tab:zero_shot_results}, \algname outperforms all baselines, though the improvements are less pronounced compared to zero-shot classification.  
This smaller margin of improvement may be attributed to the increased complexity of retrieval tasks, which require fine-grained image-text alignment across varying resolutions and object details -- challenges distinct from the more direct pattern recognition in classification.  

\subsection{Robustness to Distribution Shifts}\label{subsec:zero_shot_shifted}
To evaluate the robustness of \algname under distribution shifts, we test all approaches on three variants of ImageNetV2~\citep{recht2019imagenet} and ImageNet-R~\citep{hendrycks2021many}, which assess resilience to different real-world deviations from ImageNet1K.  
As shown in \tabref{tab:zero_shot_shifted_results}, \algname consistently outperforms all baselines, achieving the highest accuracy across all datasets.  
These results indicate that \algname not only improves standard zero-shot classification but also enhances adaptation to real-world distribution shifts.

\subsection{Linear Probing}\label{subsec:linear_probe}
We further assess whether the advantages of ranking consistency persist when supplemented with in-domain supervision.  
Specifically, we apply linear probing, where pretrained encoders remain fixed while a logistic regression classifier is trained on domain-specific datasets.  
We evaluate on 10 standard image classification benchmarks, including CIFAR-10, CIFAR-100, DTD~\citep{cimpoi14describing}, FGVG-Aircraft~\citep{maji13fine-grained}, Food101~\citep{bossard14}, GTSDB~\citep{Stallkamp2012}, OxfordPets~\citep{parkhi2012cats}, SST2~\citep{socher2013recursive}, STL-10~\citep{coates2011analysis}, and SVHN~\citep{netzer2011reading}.  
As shown in \tabref{tab:linear_probe_results}, \algname achieves the highest average accuracy, demonstrating that ranking consistency enhances generalization even with additional in-domain supervision. 

\section{Ablation Studies}\label{sec:ablation}

\subsection{Different Weights of \algname Loss}\label{subsec:component_analysis}
\begin{figure}[t]
    \centering
    \includegraphics[width=1\linewidth]{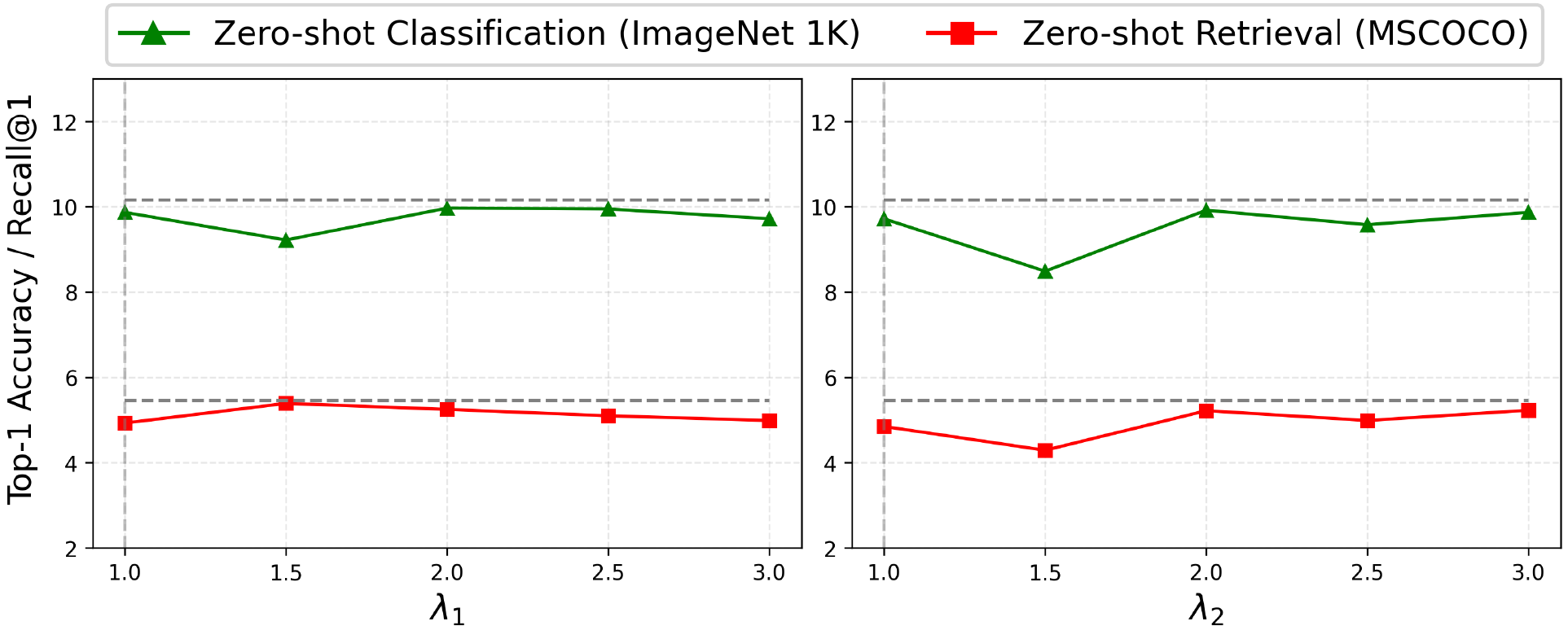}
    \vspace{-0.2in}
    \caption{
         Effect of $\lambda_1$ and $\lambda_2$ on zero-shot classification (ImageNet1K) and retrieval (MSCOCO).
    }
    \label{fig:weights_ablation}
    \vspace{-0.1in}
\end{figure}

In \eqnref{eqn:rankclip_loss}, we define the \algname loss as a linear combination of the original contrastive loss and the in-modal and cross-modal ranking consistency losses, weighted by $\lambda_1$ and $\lambda_2$, respectively. 
In \secref{subsec:training_recipe}, we introduce a training strategy that adaptively adjusts $\lambda_1$ and $\lambda_2$ at different stages of training. 
In this section, we analyze the impact of these weights and demonstrates that adaptive weighting further enhances \algname’s performance. 
Notice that all model variants follow the same pretraining setup detailed in \appref{app:training_procedure}. 
As shown in \figref{fig:weights_ablation}, \algname outperforms CLIP even with fixed $\lambda_1$ and $\lambda_2$, highlighting the effectiveness of ranking consistency. 
And the adaptive weighting strategy further boosts accuracy by preventing the under-utilization of ranking consistency at low weights and avoiding disruptions to contrastive learning at high weights.
\begin{figure*}[t]
    \centering
    \includegraphics[width=.9\linewidth]{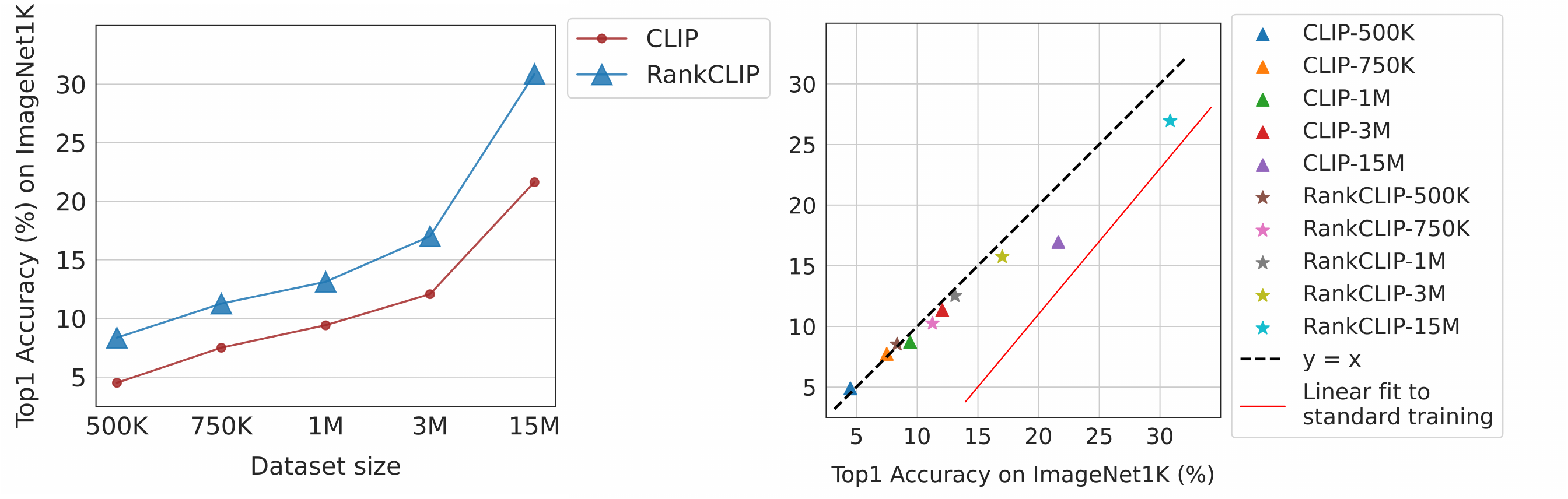}
    \vspace{-0.1in}
    \caption{
        Ablation studies of CLIP and \algname trained with different data sizes.
        \textit{Left}: zero-shot top-1 classification accuracy on ImageNet1K with various 
        data sizes randomly sampled from CC3M.
        \algname consistently outperforms CLIP with significant margins.
        \textit{Right}: zero-shot top-1 classification accuracy on ImageNet1K (horizontal axis) 
        and ImageNet1K-R (vertical axis).
        \algname demonstrates better robustness as well as accuracy.
    }
    \label{fig:data_scaling}
    % \vspace{-0.15in}
\end{figure*}
\begin{table*}[t]
\centering
\vspace{-0.1in}
\setlength{\tabcolsep}{10pt}
\renewcommand{\arraystretch}{1.2}
\resizebox{\linewidth}{!}{
\begin{tabular}{c|c|ccc|ccc|ccc|c}
\hline
\multirow{3}{*}{Method} & \multirow{3}{*}{\makecell{Vision\\Backbone}} & \multicolumn{3}{c|}{\multirow{2}{*}{ImageNet1K}} & \multicolumn{6}{c|}{MSCOCO} & \multirow{3}{*}{\makecell{Linear Probing\\Avg. Acc.}} \\
\cline{6-11}
 &  & \multicolumn{3}{c|}{} & \multicolumn{3}{c|}{Image-to-Text Retrieval} & \multicolumn{3}{c|}{Text-to-Image Retrieval} &  \\
\cline{3-11}
 &  & Top-1 & Top-3 & Top-5 & R@1 & R@5 & R@10 & R@1 & R@5 & R@10 &  \\
\hline
CLIP~\citep{radford2021learning}    & \multirow{2}{*}{RN50} & 21.6\% & 36.9\% & 44.9\% & 15.6\%  & 36.4\%  & 48.4\%  & 6.7\%   & 15.2\%  & 20.1\%  & 64.2\% \\
\cellcolor{gray!20}\algname &                      & \cellcolor{gray!20}\textbf{30.9\%} & \cellcolor{gray!20}\textbf{49.4\%} & \cellcolor{gray!20}\textbf{57.6\%} & \cellcolor{gray!20}\textbf{19.5\%}  & \cellcolor{gray!20}\textbf{42.6\%}  & \cellcolor{gray!20}\textbf{54.8\%}  & \cellcolor{gray!20}\textbf{7.5\%}   & \cellcolor{gray!20}\textbf{16.2\%}  & \cellcolor{gray!20}\textbf{21.6\%}  & \cellcolor{gray!20}\textbf{68.9\%} \\
\hline
CLIP~\citep{radford2021learning}    & \multirow{2}{*}{ViT-B/32} & 20.7\% & 35.0\% & 42.4\% & 11.9\%  & 29.4\%  & 40.8\%  & 5.1\%   & 12.9\%  & 17.9\%  & 60.7\% \\
\cellcolor{gray!20}\algname &                      & \cellcolor{gray!20}\textbf{26.2\%} & \cellcolor{gray!20}\textbf{41.4\%} & \cellcolor{gray!20}\textbf{48.9\%} & \cellcolor{gray!20}\textbf{13.8\%}  & \cellcolor{gray!20}\textbf{33.8\%}  & \cellcolor{gray!20}\textbf{45.9\%}  & \cellcolor{gray!20}\textbf{6.0\%}   & \cellcolor{gray!20}\textbf{13.6\%}  & \cellcolor{gray!20}\textbf{18.6\%}  & \cellcolor{gray!20}\textbf{61.3\%} \\
\hline
\end{tabular}
}
\caption{
    Zero-shot evaluation of CLIP and \algname trained with different vision backbones (ResNet-50 (RN50) and ViT-B/32) on ImageNet1K classification, MSCOCO cross-modal retrievals, and linear probing.
    ``R@k" denotes Recall@k.
}
\label{tab:zero_shot_retrieval_results}
\vspace{-0.15in}
\end{table*}

\subsection{Different Data Sizes}\label{subsec:scaling}
To assess the scalability of \algname, we trained both CLIP and \algname on 500k, 750k, 1M, and 3M text-image pairs from CC3M, as well as 15M pairs from YFCC15M, following the procedure in \appref{app:training_procedure}.
\figref{fig:data_scaling} compares their performance on zero-shot top-1 classification accuracy for ImageNet1K (left) and averaged linear probing results (middle), where \algname consistently outperforms CLIP.
Full, non-averaged linear probing results are provided in \appref{tab:linear_probe_ablation_full_results}.
Notably, the performance gains of \algname become more pronounced as dataset size scales from 1m to 15m, highlighting its superior scalability, which is critical for language-image pretraining.

\figref{fig:data_scaling} (right) further illustrates \algname's robustness across different dataset sizes.
The horizontal axis represents top-1 accuracy on standard ImageNet1K, while the vertical axis shows accuracy on ImageNet1K-R.
The black diagonal ($y=x$) denotes ideal robustness, with deviations below it indicating degradation under distribution shifts.
\algname remains well above the red baseline, which represents typical in-distribution to out-of-distribution generalization~\citep{miller2021accuracy}, and stays close to the ideal line, demonstrating strong robustness.

\vspace{-0.1in}

\subsection{Different Backbones: RN50 vs. ViT}\label{subsec:rn50_vs_vit}
We compare \algname and CLIP across ResNet-50 (RN50) and ViT-B/32 backbones to assess its generalization. 
As shown in \tabref{tab:zero_shot_retrieval_results}, \algname consistently outperforms CLIP across all tasks.
Specifically, for zero-shot classification on ImageNet1K, \algname improves top-1 accuracy by +9.3\% with RN50 and +5.5\% with ViT-B/32, demonstrating stronger feature learning. 
In cross-modal retrieval, \algname achieves +3.9\% in image-to-text and +0.8\% in text-to-image with RN50, with smaller but consistent gains for ViT-B/32.
\algname also improves linear probing accuracy by +4.7\% with RN50 and +0.6\% with ViT-B/32, confirming its advantage in representation learning. 
While both architectures benefit, RN50 sees the largest gains, suggesting that ranking consistency particularly enhances hierarchical feature extraction in CNNs.
%
% \begin{figure}[t]
%     \centering
%     \begin{subfigure}[b]{0.48\linewidth} % 第一行左图
%         \includegraphics[width=\linewidth]{figures/linear_probing_datasize_3m_RN.png}
%         \caption{
%             ResNet 50
%         }
%         \label{fig:data_scaling_1}
%     \end{subfigure}
%     \hfill
%     \begin{subfigure}[b]{0.48\linewidth} % 第一行右图
%         \includegraphics[width=\linewidth]{figures/linear_probing_datasize_3m_ViT.png}
%         \caption{
%             ViT-B/32
%         }
%         \label{fig:data_scaling_2}
%     \end{subfigure}
    
%     \vspace{0.1in} % 调整两行之间的间距
    
%     \begin{subfigure}[b]{0.48\linewidth} % 第二行左图
%         \includegraphics[width=\linewidth]{figures/linear_probing_datasize_15m_RN.png}
%         \caption{
%             ResNet 50
%         }
%         \label{fig:data_scaling_3}
%     \end{subfigure}
%     \hfill
%     \begin{subfigure}[b]{0.48\linewidth} % 第二行右图
%         \includegraphics[width=\linewidth]{figures/linear_probing_datasize_15m_ViT.png}
%         \caption{
%             ViT-B/32
%         }
%         \label{fig:data_scaling_4}
%     \end{subfigure}
    
%     % \vspace{-0.2in} % 调整 caption 与图片的间距
%     \caption{
%         Ablation studies of CLIP and \algname trained with different datasize and basemodel.
%         %
%         (a) and (b) show the performance on 3M dataset.
%         %
%         (c) and (d) show the performance on 15M dataset.
%     }
%     \label{fig:data_scaling}
% \end{figure}

\begin{figure*}[ht]
    \centering
    \includegraphics[width=.9\linewidth]{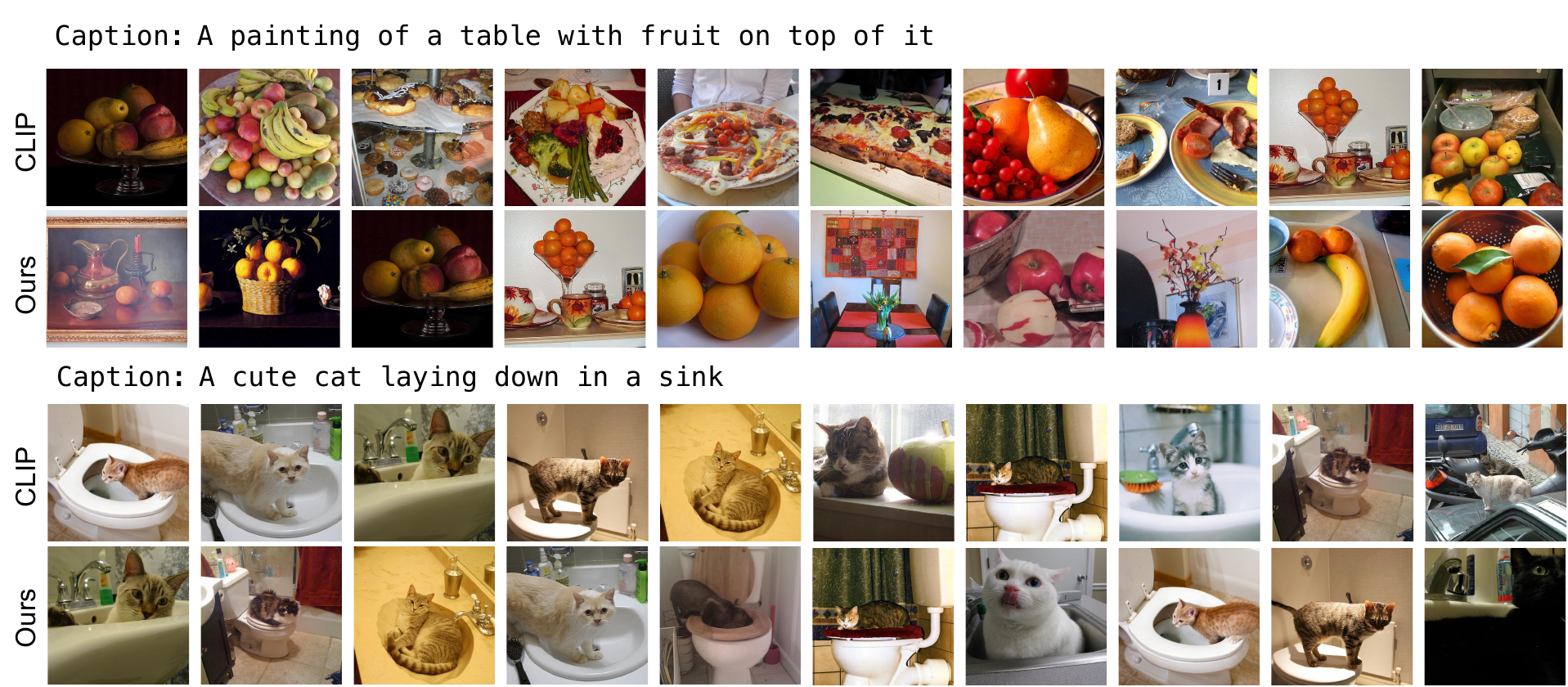}
    \vspace{-0.1in}
    \caption{
        For a given text query, we present the top ten most semantically relevant images (ordered from left to right) obtained through both CLIP and \algname. In comparison to CLIP, our approach consistently retrieves images that more comprehensively align with the textual description, maintaining this advantage even after the correct reference image appears in the ranked results.
    }
    \label{fig:t2i_examples}
    \vspace{-0.1in}
\end{figure*}
\section{Analysis}\label{sec:analysis}
\subsection{Modality Gap}\label{subsec:modality_gap}
\begin{figure}[ht]
    \centering
    \includegraphics[width=.9\linewidth]{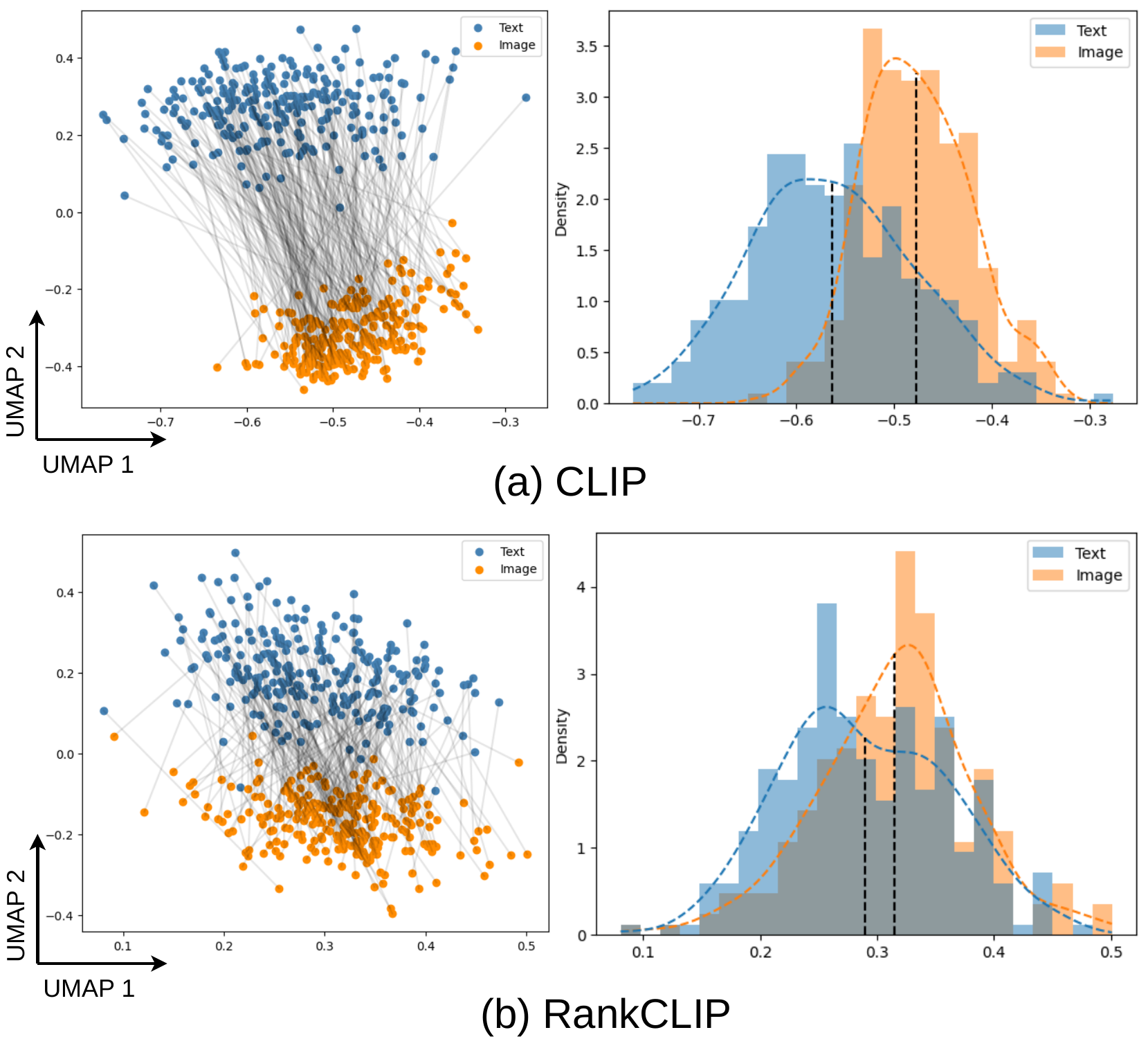}
    \vspace{-0.1in}
    \caption{
        Scatter and histograms plots illustrating modality gaps of (a) CLIP and (b) \algname.
    }
    \label{fig:modality_gap}
    % \vspace{-0.15in}
\end{figure}

We analyze the modality gaps of CLIP and \algname by visualizing 250 text-image pair embeddings, reduced to two dimensions using UMAP~\citep{mcinnes2018umap}, and presenting a histogram of the gaps.
The modality gap~\citep{liang2022mind} refers to the separation between text and image embeddings in multimodal models, hindering joint representation learning. 
This gap, inherent from initialization and reinforced by contrastive learning in CLIP, challenges effective language-image modeling.
Recent studies~\citep{srivastava2024omnivec, kumar2024improving, oh2024geodesic} suggest that reducing this gap improves multimodal representations and downstream performance.
As shown in \figref{fig:modality_gap}, \algname exhibits a significantly smaller modality gap than CLIP, demonstrating that our ranking consistency approach effectively enhances text-image alignment.

\subsection{Alignment and Uniformity}\label{subsec:uniformity}
Beyond reducing modality gap, effective contrastive learning should ensure a broad and uniform distribution over a hypersphere~\citep{wang2020understanding}. 
These objectives -- similarity and uniformity -- are quantified by alignment and uniformity scores, respectively.
Following~\citet{goel2022cyclip} and the notations in \secref{sec:rankclip}, we compute the alignment score $S_{\text{A}}$ and uniformity score $S_{\text{U}}$ as:
\begin{align}
    S_{\text{A}} &= \frac{1}{N}\sum_{j=1}^N \hat{I}^T_j \hat{T}_j, \\
    S_{\text{U}} &= \log \left( \frac{1}{N(N-1)}\sum_{j-1}^N \sum_{k=1, j \neq k}^N \exp^{-\hat{I}^T_j \hat{T}_k} \right)
\end{align}
where $N$ is the number of text-image pairs.
$S_{\text{A}}$ captures the average cosine similarity between corresponding text and image embeddings, while $S_{\text{U}}$ measures how evenly embeddings are spread across the space. 
High alignment indicates strong correlation between paired embeddings, whereas low uniformity suggests diverse and efficient embedding distribution—desirable for tasks like retrieval.
As shown in \tabref{tab:alignment_uniformity_results}, CLIP achieves stronger alignment but suffers from poor uniformity, leading to redundant representations. 
On the other hand, \algname along with two of its ablated version, $\textsc{RankCLIP}_I$ and $\textsc{RankCLIP}_C$ presents much better balance between alignment and uniformity.
These results further suggest that optimizing solely for alignment or uniformity does not necessarily translate to better task performance.
\begin{table}[t]

\centering
\setlength{\tabcolsep}{8pt}
\renewcommand{\arraystretch}{1.2}
\resizebox{\linewidth}{!}{
\begin{tabular}{c | c c | c c | c c }
\hline
& \multicolumn{2}{c|}{CIFAR-10} & \multicolumn{2}{c|}{CIFAR-100} & \multicolumn{2}{c}{ImageNet1K} \\
\hline
& $S_{\text{A}}$ & $S_{\text{U}}$  & $S_{\text{A}}$ & $S_{\text{U}}$ & $S_{\text{A}}$ & $S_{\text{U}}$  \\
\hline
CLIP & \textbf{0.28} & -0.19 & \textbf{0.28} & -0.18  & \textbf{0.33} & -0.19   \\
\algname  & 0.23 & -0.14  & 0.26 & -0.13  & 0.29 & \textbf{-0.13} \\
$\textsc{RankCLIP}_C$ & 0.23  & \textbf{-0.13}  & 0.24 & \textbf{-0.12} & 0.29 & -0.14  \\
$\textsc{RankCLIP}_I$ & 0.25 & -0.15  & 0.28  & -0.14  & 0.32 & -0.17  \\
\hline
\end{tabular}
}
\caption{
    Alignment and uniformity scores of CLIP, \algname. $\textsc{RankCLIP}_C$ and $\textsc{RankCLIP}_I$ indicate the solely cross-modal and in-modal rank loss.
}
\label{tab:alignment_uniformity_results}
\vspace{-0.2in}
\end{table}

\begin{figure*}[ht]
    \centering
    \includegraphics[width=.9\linewidth]{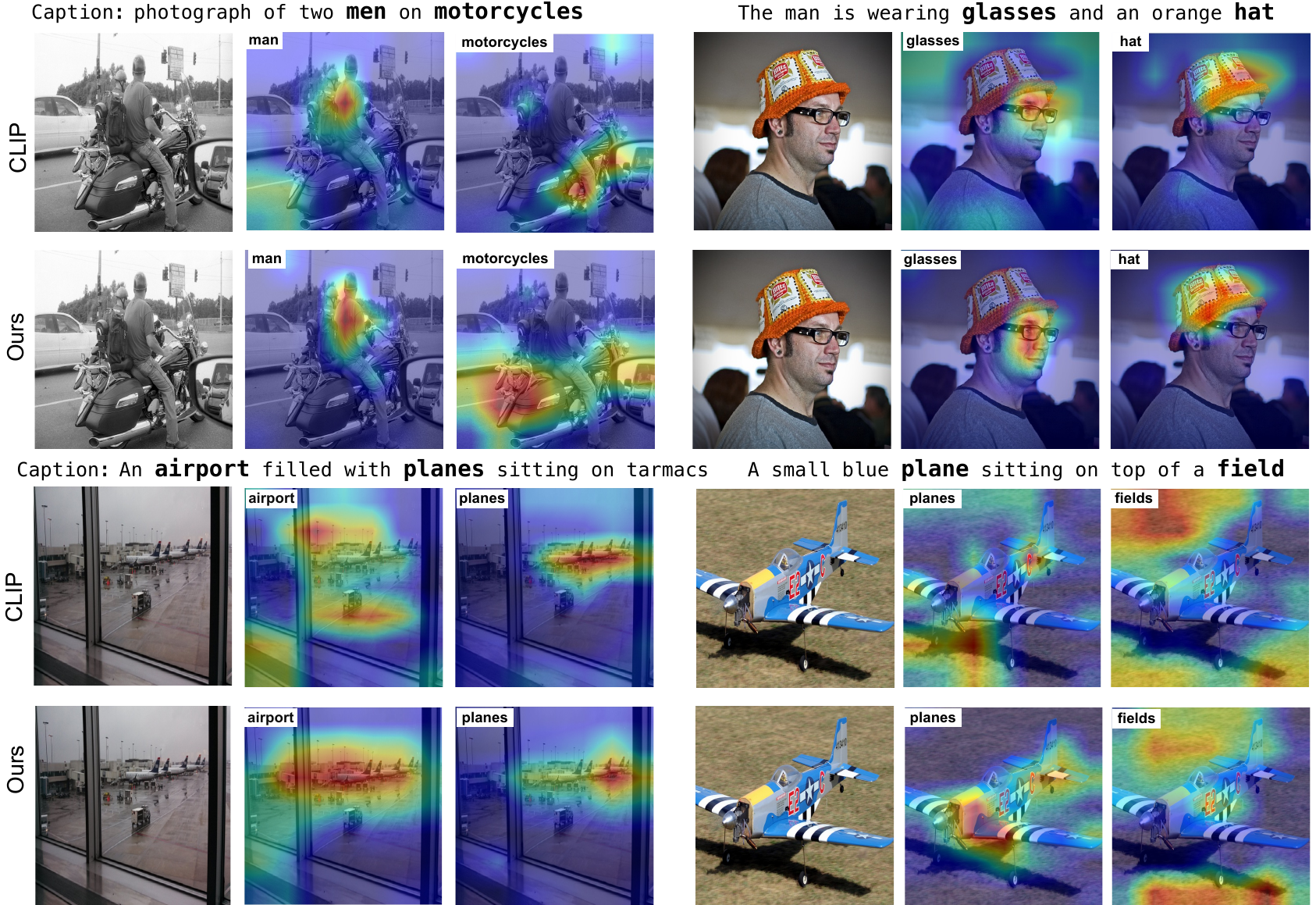}
    \caption{
        Class activation maps for \algname and CLIP on different objects in the caption from MSCOCO. \algname has more precise responses to some nouns compared to CLIP and can accurately locate the region related to the noun.
    }
    \label{fig:grad_cam}
    % \vspace{-0.15in}
\end{figure*}

% \vspace{-0.2in}
\subsection{Qualitative Examples}\label{subsec:qualitive_examples}
\paragraph{Class activation maps.}

To further examine the effects of ranking consistency, we visualize class activation maps (CAMs)~\citep{selvaraju2017grad} for \algname and CLIP in \figref{fig:grad_cam}. 
The results show that \algname consistently attends to more semantically relevant regions in the images. 
For example, when given the caption \texttt{`An airport filled with planes sitting on tarmacs'}, CLIP mistakenly highlights surrounding areas, whereas \algname focuses precisely on the planes.
Similar improvements are observed across other examples, demonstrating that \algname better aligns textual descriptions with visual concepts. 
This suggests that ranking consistency enhances fine-grained feature learning, leading to more localized and accurate visual grounding.

\paragraph{Text-to-image retrieval.}

\figref{fig:t2i_examples} compares text-to-image retrieval results from \algname and CLIP. 
Given a query, we display the top-ranked images retrieved by each model. 
\algname consistently retrieves more semantically aligned images, even beyond the correct reference image. 
For instance, in the example of \texttt{'A cute cat laying down in a sink'}, correctly identifies a cat in a sink, whereas CLIP misidentifies it due to the visual similarity between sinks and toilets. 
This demonstrates \algname’s ability to capture fine-grained semantic distinctions, reinforcing its advantage in retrieval tasks that demand precise understanding.
\section{Conclusion}\label{sec:conclusion}
In this paper, we introduce \algname, a novel language-image pretraining method that integrates 
ranking consistency into the contrastive learning paradigm. 
\algname aims to better understand the complex many-to-many relationships in diverse text-image 
pairs by optimizing a self-supervised, list-wise rank loss. 
Through extensive experiments, including zero-shot classification, robustness to distribution 
shifts, linear probing, and zero-shot image-text retrieval, \algname not only enhances 
performance but also improves model robustness and semantic comprehension, outperforming the 
baseline CLIP and another state-of-the-art model ALIP.
Our ablation studies and analyses further demonstrate and interpret the significance of each 
component of \algname in boosting performance and understanding across modalities. 
We believe that the methodologies and principles of \algname will inspire further research and 
lead to the development of models with a deeper understanding of the intricate interactions 
between visual and textual data.

\clearpage
{
    \small
    \bibliographystyle{ieeenat_fullname}
    \bibliography{custom}
}

\clearpage
\clearpage
\appendix
\onecolumn

\section*{Appendix}
\section{Training Procedures}\label{app:training_procedure}
\subsection{Implementation Details}\label{app:implementation}
For CLIP~\citep{radford2021learning}, we use the official implementation released by
OpenAI\footnote{CLIP repository on GitHub: https://github.com/openai/CLIP.}.
And for ALIP~\citep{yang2023alip}, we also use the official implementation released by
the paper authors\footnote{ALIP repository on GitHub: https://github.com/deepglint/ALIP.}.
As the proposed \algname essentially shares the same model architecture (separate vision, text
encoders, projection layer, and a classification head) as CLIP, we build upon the CLIP code 
repository for our model construction\footnote{\algname repository will be released upon 
acceptance.}.
We set the scaling parameters for cross-modal ($\lambda_{c}$) and in-modal ($\lambda_{i}$) 
ranking consistency to 1/16 and 1/16 respectively throughout all the experiments unless 
otherwise noted.
All CLIP, ALIP and \algname models are initialized from scratch without loading any 
existing weights.
And the embedding sizes for both modalities all project to 1024 across the three models.

\subsection{Training Parameters}\label{app:training_parameters}
Following CLIP~\citep{radford2021learning}, we adopt the 
ResNet-50~\citep{he2016deep} and transformer architectures~\citep{devlin2018bert} for image 
and text encoding, respectively. 
Training is conducted from scratch over 64 epochs using a single NVIDIA A100 GPU, with a batch 
size of 512, an initial learning rate of 0.0005 employing cosine scheduling, and 10,000 
warm-up steps. 

\subsection{Training Time Consumption}\label{app:time_consumption}
we conducted the experiments using the same hardware specifications. The table below shows the time consumption for training our RankCLIP and CLIP models with 50K samples from CC3M using a single NVIDIA A100 GPU.
\begin{table}[htbp]
    \centering
    \caption{Training Details}
    \begin{tabular}{|c|c|c|c|c|c|}
        \hline
        & Time consumption& Dataset size & epochs & batch\_size & model\_name \\
        \hline
        CLIP & 1d 2h 54m 48s & 50K & 64 & 512 & RN50 \\
        \hline
        \algname & 1d 1h 4m 23s & 50K & 64 & 512 & RN50 \\
        \hline
    \end{tabular}
    \label{tab:training_details}
\end{table}

As shown in the table, the difference in time consumption is negligible. Interestingly, our method is slightly faster than CLIP, but we think it may be attributed to hardware optimizations or variance.

\section{CLIP Preliminaries}\label{app:clip}
CLIP~\citep{radford2021learning} has been a prominent method for learning detailed multimodal 
representations through the alignment of images and texts. 
Given a set $\mathcal{D} = \{(V_j, T_j)\}_{j=1}^{N}$ of $N$ image-text pairs, where 
$V_j$ denotes an image and $T_j$ is the corresponding text, the goal is to learn 
representations that map semantically similar images and texts closer in the embedding space, 
while dissimilar pairs are distanced apart.
More specifically, the foundational CLIP model employs two encoders: an image encoder 
$f_I: \mathcal{I} \rightarrow \mathbb{R}^m$ that processes raw images into visual embeddings 
and a text encoder $f_T: \mathcal{T} \rightarrow \mathbb{R}^n$ which encodes textual data 
into text embeddings.
Then both the text and visual features are projected to a latent space with identical dimension.
Formally, the embeddings for a text-image pair $(V_j, T_j)$ are denoted as $v_k = f_I(V_j)$ 
and $t_j = f_T(T_j)$, respectively.
The embeddings are then normalized to lie on an unit hypersphere by enforcing 
$l_2$-norm constraint:
\begin{equation}
\hat{v}_j = \frac{v_j}{\|v_j\|_2}, \quad \hat{t}_j = \frac{t_j}{\|t_j\|_2}.
\end{equation}
so that the magnitude information is erased and only direction is preserved.

To align the image and text representations, a contrastive loss function, typically a 
variant of the InfoNCE loss~\citep{oord2018representation}, which optimizes the similarity of the matched pair against unmatched pairs, is utilized, i.e.:
\begin{align}
% \scriptsize
\mathcal{L}_{\text{CLIP}} = -\frac{1}{2N} \sum_{j=1}^{N} \biggr[ 
\underbrace{\log \frac{\exp(\hat{v}_j^\top \hat{t}_j / \tau)}{\sum_{k=1}^{N} 
\exp(\hat{v}_j^\top \hat{t}_k / \tau)}}_{{\color{red} \circled{1}}}
+ 
\underbrace{\log \frac{\exp(\hat{t}_j^\top \hat{v}_j / \tau)}{\sum_{k=1}^{N} 
\exp(\hat{t}_j^\top \hat{v}_k / \tau)}}_{{\color{blue} \circled{2}}}
\biggr]
\label{eqn:clip loss}
\end{align}
where the first term ${\color{red} \circled{1}}$ contrasts images with the texts, 
the second term ${\color{blue} \circled{2}}$ contrasts texts with the images, 
and $\tau$ denotes a temperature scaling parameter that adjusts the concentration 
of the distribution.
The optimization of Eqn.~(\ref{eqn:clip loss}) results in embeddings where the cosine similarity 
between matched image-text pairs is maximized in comparison to unmatched pairs, thus 
achieving the desired alignment in the joint embedding space.

Despite the efficacy of CLIP in learning correlated multimodal embeddings, it inherently 
relies on strict pairwise matched comparisons and fails to capture the more complex, 
fine-grained nature of semantic similarity within and across modalities that are 
generally treated as unmatched. 
This observation motivates the development of \algname, which innovates beyond binary pairwise 
contrasts to consider holistic listwise consistency within and across modalities.

\subsection{Additional Experiments}
We conduct the linear probing experiment under different training datasize from 3m to 15m as shown in 
\ref{tab:linear_probe_ablation_full_results}. 
\vspace{-0.1in}
\begin{table*}[t]
    \centering
    
\setlength{\tabcolsep}{10pt}
\renewcommand{\arraystretch}{1.2}
\resizebox{\linewidth}{!}{
\begin{tabular}{p{1cm}<{\centering}|p{2cm}<{\centering}|p{2cm}<{\centering}|c c c c c c c c c c}
\hline
\makecell[c]{\textbf{Data} \\ \textbf{Size}}& \makecell[c]{\textbf{Method}}&\makecell[c]{\textbf{Model} \\ \textbf{Type}} & \rotatebox{65}{CIFAR-10} & \rotatebox{65}{CIFAR-100} & \rotatebox{65}{DTD} & \rotatebox{65}{FGVGAircraft} & \rotatebox{65}{Food101} & \rotatebox{65}{GTSRB}  & \rotatebox{65}{OxfordPets} & \rotatebox{65}{SST2} & \rotatebox{65}{STL10} & \rotatebox{65}{SVHN}  \\
\hline
 & CLIP & RN50 & 80.12\% & 58.50\% & 57.18\% & 39.75\% & 59.14\% & 72.41\% & 61.73\% & 54.48\% & 86.01\% & 58.92\%       \\
3m & \algname & RN50 & 78.29\% & 56.24\% & 57.82\% & 39.30\% & 58.63\% & 74.13\% & 64.35\% & 55.02\% & 86.69\% & 60.68\%     \\
 & CLIP & ViT-B/32 & 77.60\% & 56.15\% & 43.19\% & 22.59\% & 39.72\% & 62.05\% & 40.39\% & 50.96\% & 78.99\% & 50.53\%    \\
 & \algname & ViT-B/32 & 78.42\% & 56.64\% & 42.39\% & 23.43\% & 40.19\% & 60.63\% & 40.56\% & 53.32\% & 79.60\% & 47.72\%     \\
\hline
 & CLIP & RN50 & 78.81\% & 56.32\% & 61.49\% & 25.83\% & 61.64\% & 68.76\% & 60.37\% & 55.57\% & 89.82\% & 47.99\%  \\
15m & \algname & RN50 & 83.27\% & 62.96\% & 65.96\% & 32.19\% & 68.11\% & 74.25\% & 67.40\% & 56.34\% & 94.20\% & 53.06\%     \\
 & CLIP & ViT-B/32 & 82.97\% & 62.55\% & 49.47\% & 24.48\% & 52.46\% & 63.55\% & 50.78\% & 52.66\% & 87.14\% & 46.38\%     \\
 & \algname & ViT-B/32 & 82.79\% & 59.89\% & 52.50\% & 23.94\% & 56.44\% & 61.58\% & 52.98\% & 53.60\% & 89.01\% & 42.16\%     \\
\hline
\end{tabular}
}
    \caption{
        Linear probing accuracy on 10 downstream datasets. 
    }
\vspace{-0.2in}
\label{tab:linear_probe_ablation_full_results}
\end{table*}

\subsection{Pseudo-code}\label{app:Pseudo-code}
\lstset{ %
    language=Python,                % the language of the code
    basicstyle=\ttfamily\footnotesize, % the size of the fonts that are used for the code
    keywordstyle=\color{blue},       % keyword style
    commentstyle=\color{comment},       % comment style
    morekeywords={self},             % if you want to add more keywords to the set
    breaklines=true,                 % automatic line breaking only within margins
}
\vspace{-0.1in}

\begin{algorithm}[H]
\caption{Pseudo-code of \algname loss in a Python-like style.}
\begin{lstlisting}
# emb_pred: predictions from the model, shape [embs_length, embs_length]
# emb_true: ground truth labels, shape [embs_length, embs_length]

def rank_loss(emb_pred, emb_true):
    # Shuffle for randomised tie resolution
    emb_pred_shuff = emb_pred[:, random_indices]
    emb_true_shuff = emb_true[:, random_indices]
    # Record the rank label index
    emb_true_sorted, indices = emb_true_shuff.sort(descending=True, dim=-1)
    # Ranking the pred embedding by the true indices
    preds_sorted = gather(emb_pred_shuff, dim=1, index=indices)
    # Implementation of the Eq.1, Eq.2 and Eq.3
    max_pred_values, _ = preds_sorted.max(dim=1, keepdim=True)
    preds_sorted_minus_max = preds_sorted - max_pred_values
    cumsums = cumsum(preds_sorted_minus_max.exp().flip(dims=[1]), dim=1).flip(dims=[1])
    loss = (log(cumsums) - preds_sorted_minus_max) * scale_factor
    return mean(sum(loss, dim=1))

# Cross-modal embeddings
logits_text_per_image=image_embeds @ text_embeds.T
logits_iamge_per_text=logits_text_per_image.T
# In-modal embeddings
logits_image_per_image=image_embeds @ image_embeds.T 
logits_text_per_text=text_embeds @ text_embeds.T
# Compute the cross-modal rank loss
Cross_modal_loss=rank_loss(logits_text_per_image,logits_image_per_text)+rank_loss(logits_image_per_text, logits_text_per_image)
# Compute the in-modal rank loss
In_modal_loss=rank_loss(logits_image_per_image,logits_text_per_text)+rank_loss(logits_text_per_text, logits_image_per_image)
# Rank loss
Rank_loss=Contrastive_loss+Cross_modal_loss+In_modal_loss
\end{lstlisting}
\label{alg:pseudocode}
\end{algorithm}
\vspace{-0.1in}

\end{document}